\documentclass[11pt]{article}

\usepackage[preprint]{acl}

\usepackage{times}
\usepackage{latexsym}

\usepackage[T1]{fontenc}

\usepackage[utf8]{inputenc}

\usepackage{microtype}

\usepackage{inconsolata}

\usepackage{graphicx}

\usepackage{amsmath}
\usepackage{amssymb}
\usepackage{mathtools}
\usepackage{amsthm}
\usepackage[inline]{enumitem}
\usepackage{multirow}
\usepackage{dblfloatfix}
\usepackage{algorithm}    
\usepackage{algorithmic} 
\usepackage{booktabs}
\usepackage{caption}    
\usepackage{subcaption} 

\title{Higher-Dimensional Rotary Position Embedding}

\author{\textbf{Yixing Li\textsuperscript{2}$^\dagger$, Ruobing Xie\textsuperscript{1}$^\ast$, Yudong Zhang\textsuperscript{3}, Yushi Bai\textsuperscript{1}, Xingwu Sun\textsuperscript{1,4}, Yu Cheng\textsuperscript{2}$^\ast$}  
\\
 \textsuperscript{1}Tencent Hunyuan 
 \textsuperscript{2}The Chinese University of Hong Kong 
 \\
 \textsuperscript{3}Tsinghua University 
 \textsuperscript{4}University of Macau
 \\
\texttt{li.yixing@outlook.com}~~~
\texttt{xrbsnowing@163.com}~~~
\texttt{chengyu@cse.cuhk.edu.hk}
}

\begin{document}
\maketitle
\begin{abstract}

Transformers rely on position embedding mechanisms in long context modeling in most cases. Rotary Position Embedding (RoPE) embeds positional information with independent 2D rotations, forming relative position terms in self-attention. However, its pairwise, block-based, and decoupled structure limits deep mixing and robustness across channels.
We propose HD-RoPE, which extends RoPE from independent 2D rotations to higher-dimensional rotations and introduces a Paley-I orthogonal basis to obtain balanced, isotropic, and dense phase mixing within each rotation subspace.
This significantly enhances channel coupling and rotational degrees of freedom while maintaining orthogonal stability and the relative position closure property. Furthermore, HD-RoPE is easily optimized for engineering efficiency without introducing additional trainable parameters.
We have conducted extensive evaluation results demonstrating that HD-RoPE achieves significant performance improvements over standard RoPE across various popular benchmarks and in both long and short contexts.
\end{abstract}

\section{Introduction}
\let\thefootnote\relax\footnotetext{$^\ast$ Corresponding author.}
\let\thefootnote\relax\footnotetext{$^\dagger$ Work conducted during internship at Tencent.}

\begin{figure}[!t]
  \begin{center}
    \centerline{\includegraphics[width=0.95\columnwidth]{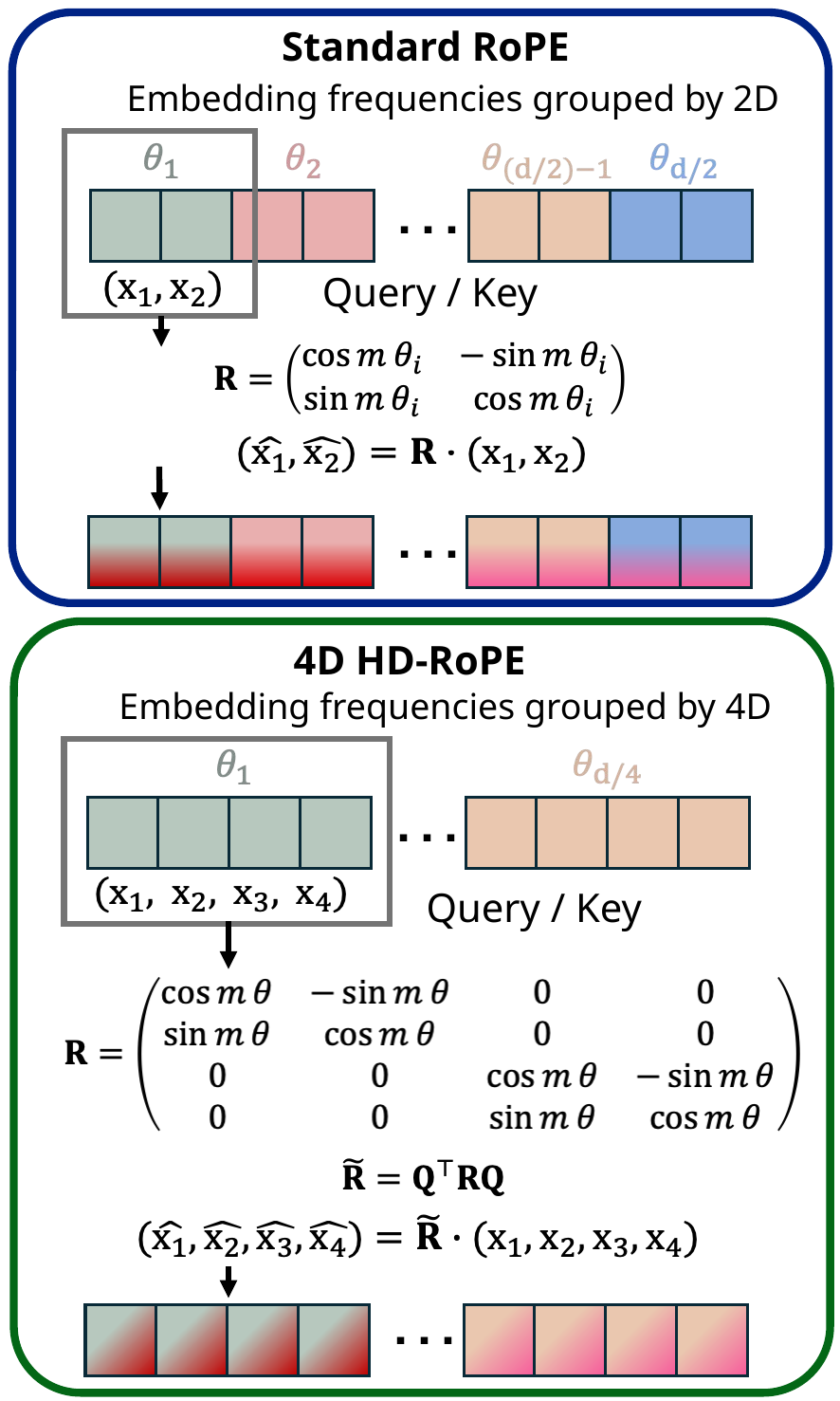}}
    \caption{The illustration of our HD-RoPE. Q denotes the selected orthogonal matrix in Section \ref{sec:method:main}, enabling more isotropic, balanced, and dense phase interaction.}
    \label{fig:main1}
  \end{center}
\end{figure}





Transformer architecture has established itself as the dominant backbone for large language models (LLMs) across a wide range of domains. They heavily rely on efficient positional encoding mechanisms in long sequence modeling. Rotary Position Embedding (RoPE; \citealp{su2024roformer}), because it encodes absolute position as rotation and explicitly forms relative position signals in self-attention, has become one of the default choices for current mainstream language models.

When models need to extrapolate from the pre-training context length to longer contexts, RoPE exhibits typical degradation.
Hence, a large body of effort on RoPE has focused on modifying the frequency schedule or angular velocity, such as rescaling the RoPE base, interpolating positions, or selectively weakening positional bias \cite{chen2023extending, peng2023yarn, ding2024longrope, yang2025rope}. These methods improve how phases evolve across positions. 

However, a more fundamental assumption of RoPE that position is injected only through several mutually independent two-dimensional rotations has largely remained untouched.
RoPE can be roughly understood as: projecting the representation vector of each token onto several two-dimensional planes, and rotating it within each plane by a position-dependent angle (phase). The physical meaning of two-dimensional rotation is very intuitive and sound, while it also indicates that standard RoPE binds each positional frequency to an isolated two-dimensional channel pair. In this case, each relative phase can interact with only two fixed-dimension Q/K features, making the phase-sensitive attention term sparse and pairwise.

Based on the above intuition, this paper proposes \textbf{HD-RoPE}, a method for injecting positional information that generalizes the two-dimensional RoPE system to an higher $N$-dimensional rotation space. HD-RoPE groups channels into sets of N, applying rotations within $N$-dimensional subspaces. 
This changes the rotation unit of RoPE from an isolated 2D pair to a higher-dimensional phase subspace, allowing the same positional phase to be shared and mixed across more Q/K channels. 
Within these subspaces, we instantiate the mixing basis using Paley-I conference matrices \cite{Paley1933OnOM} to obtain a \emph{\textbf{more isotropic, balanced, and dense phase interaction}}, enabling positional phases to couple more sufficiently with Q/K channels under the basic invariants of RoPE, which is effective, efficient and parameter-free. 

This significantly enhances channel mixing and rotational degrees of freedom while maintaining the stability and norm preservation of orthogonal transformations, further enhancing the relative position encoding capability of RoPE. 
The main contributions of HD-RoPE include:
\begin{enumerate}
    \item We propose a higher-dimensional rotational perspective, 
    providing a general framework for high-dimensional scaling, wherein 4D represents the sweet spot for balancing performance and efficiency. 
    
    \item We adopt the conference matrix as a fixed and structurally simple orthogonal mixing transformation (balanced, low-coherence, and parameter-free) to enhance better cross-channel coupling without introducing additional trainable parameters.
    \item We conducted extensive experiments and found that HD-RoPE significantly outperforms the original RoPE in various tasks with various settings, while maintaining almost no additional computational overhead.
\end{enumerate}


\section{Preliminary}
\paragraph{Self-Attention}
Given an input sequence of length $N$, its embedding without positional information is denoted as $\mathbf{E_N}=\{\mathbf x_i\}_{i=1}^{N}$, where $\mathbf{x_i}\in\mathbb R^{d}$. The self-attention of Transformer maps the representation of each position to:
\begin{equation}
\mathbf q_m=f_q(\mathbf x_m,m),\,
\mathbf k_n=f_k(\mathbf x_n,n).
\end{equation}
The output is then calculated as:
\begin{equation}
\begin{aligned}
\mathbf a_{m,n}&=
\frac{\exp\!\left(\mathbf q_m^{\top}\mathbf k_n/\sqrt{d}\right)}
{\sum_{j=1}^{N}\exp\!\left(\mathbf q_m^{\top}\mathbf k_j/\sqrt{d}\right)},\\
\,\,
\mathbf o_m&=\sum_{n=1}^{N}a_{m,n}\mathbf v_n.
\end{aligned}
\end{equation}
The essential function of positional encoding is to design $f_q,f_k,f_v$ or their equivalent implementations so that the inner product $\mathbf q_m^{\top}\mathbf k_n$ explicitly carries the positional signal in the attention weights.

\paragraph{Rotary Position Embedding (RoPE)}
RoPE starts with the relative position objective and imposes structured constraints on the attention inner product. It aims to find a function $g$ such that:
\begin{equation}
\langle f_q(\mathbf x_m,m),\,f_k(\mathbf x_n,n)\rangle
= g(\mathbf x_m,\mathbf x_n,m-n).
\end{equation}
In this form, position is injected into the attention only in the relative form $(m-n)$. RoPE provides a multiplicative implementation: the linearly projected vector is rotated in terms of position in several two-dimensional subspaces, thereby injecting a controllable relative phase difference into the inner product without changing the vector norm.

Specifically, when $d$ is even, $\mathbb R^d$ is divided into $d/2$ two-dimensional subspaces and a block diagonal rotation matrix is defined:
\begin{equation}
\begin{gathered}
\mathbf R^{d}_{\Theta,m}=
\mathrm{diag}\,\big(\mathbf R_{\theta_1,m}^{(2)},\mathbf R_{\theta_2,m}^{(2)},\ldots,\mathbf R_{\theta_{d/2},m}^{(2)}\big),\\
\mathbf R_{\theta_i,m}^{(2)}=
\begin{pmatrix}
\cos m\theta_i&-\sin m\theta_i \\
\sin m\theta_i& \cos m\theta_i 
\end{pmatrix}.
\end{gathered}
\end{equation}
Moreover, $f$ is defined as:
\begin{equation}
f_{\{q,k\}}(\mathbf x_m,m)=\mathbf R^{d}_{\Theta,m}\mathbf W_{\{q,k\}}\mathbf x_m.
\end{equation}
Orthogonality $\mathbf R_{\Theta,m}^{\top}\mathbf R_{\Theta,m}=\mathbf I$ guarantees the numerical stability and norm preservation of the encoding process: for any vector $\mathbf z$, $\|\mathbf R_{\Theta,m}\mathbf z\|_2=\|\mathbf z\|_2$. 
Rotation does not change the magnitude of the features, thus possessing numerical stability. More importantly, it produces an explicit relative positional form in the attention inner product:
\begin{equation}
\begin{aligned}
\mathbf q_m^{\top}\mathbf k_n &=
(\mathbf R^{d}_{\Theta,m}\mathbf W_q\mathbf x_m)^{\top}
(\mathbf R^{d}_{\Theta,n}\mathbf W_k\mathbf x_n) \\
&=\mathbf x_m^{\top}\mathbf W_q^{\top}\mathbf R^{d}_{\Theta,n-m}\mathbf W_k\mathbf x_n,
\end{aligned}
\end{equation}
where $\mathbf R^{d}_{\Theta,n-m}=(\mathbf R^{d}_{\Theta,m})^{\top}\mathbf R^{d}_{\Theta,n}$ depends only on the relative displacement $(n-m)$. In practice, a set of multi-scale frequencies $\Theta=\{\theta_i\}$ is often chosen (e.g., covering different frequency bands according to an exponential law) so that the model can simultaneously characterize short-range and long-range relative position signals.



\section{Method}\label{sec:method:main}

\subsection{HD-RoPE's Orthogonal conjugation preserves RoPE invariants} 

HD-RoPE aims to introduce higher-dimensional rotation for positional modeling.
Specifically, we employ orthogonal conjugation to reparameterize the rotation matrix by orientation to introduce cross-dimensional mixing. Given any orthogonal matrix $\mathbf Q$ with $\mathbf Q^{\top}\mathbf Q = \mathbf I$, we define:
\begin{equation}
\widetilde{\mathbf R}_{\Theta,m}\triangleq \mathbf Q^{\top}\mathbf R_{\Theta,m}\mathbf Q.
\label{eq.R_HD_RoPE}
\end{equation}
Based on orthogonality and conjugate structure:
\begin{equation}
\begin{aligned}
    \widetilde{\mathbf R}_{\Theta,m}^{\top}\widetilde{\mathbf R}_{\Theta,n} &=\mathbf Q^{\top}\mathbf R_{\Theta,m}^{\top}\mathbf R_{\Theta,n}\mathbf Q \\&=\mathbf Q^{\top}\mathbf R_{\Theta,n-m}\mathbf Q.
\end{aligned}
\end{equation}
Then $\widetilde{\mathbf R}_{\Theta,m}$ remains orthogonal, retaining the two critical properties:
(a). Norm Preservation: $\|\widetilde{\mathbf R}_{\Theta,m}\mathbf z\|_2=\|\mathbf z\|_2$; 
(b). Distance Preservation: $
\widetilde{\mathbf R}_{\Theta,n-m}=\widetilde{\mathbf R}_{\Theta,m}^{\top}\widetilde{\mathbf R}_{\Theta,n}.
$ 
Therefore, while \textbf{\emph{preserving relative displacement encoding and stable rotation of the original RoPE}}, $\mathbf Q$ allows us to change the coordinate system in which the rotation is applied, \textbf{\emph{increasing higher-cross-dimensional mixing and rotational degrees of freedom}}.

\begin{algorithm}[!t]
  \caption{RoPE (2D Rotation)}
  \label{alg:rope-2d-freqs}
  \begin{algorithmic}
    \STATE \textbf{Input:} embedding $\mathbf t$,  rotation frequencies $\mathbf{freqs}$
    \vspace{-1.2em}
    \STATE $\cos \leftarrow \cos(\mathrm{freqs})$;\quad $\sin \leftarrow \sin(\mathrm{freqs})$
    \STATE Split $\mathbf t$: $(\mathbf x_1,\mathbf x_2) \leftarrow \mathrm{Chunk}(\mathbf t, 2)$
    \STATE $\mathrm{RotateHalf}(\mathbf t) \leftarrow \mathrm{Concat}(-\mathbf x_2,\ \mathbf x_1)$
    \STATE $\mathbf t \leftarrow \mathbf t \odot \cos\ +\ \mathrm{RotateHalf}(\mathbf t)\odot \sin$
    \STATE \textbf{Output:} updated embedding $\mathbf t$
  \end{algorithmic}
\end{algorithm}

\begin{algorithm}[!t]
  \caption{HD-RoPE (4D Rotation; Paley-I)}
  \label{alg:hdrope-4d-paleyi-freqs}
  \begin{algorithmic}
    \STATE \textbf{Input:} embedding $\mathbf t$,  rotation frequencies $\mathbf{freqs}$
    \vspace{-1.2em}
    \STATE $\cos \leftarrow \cos(\mathrm{freqs})$;\quad $\sin \leftarrow \sin(\mathrm{freqs})$
    \STATE Split $\mathbf t$: $(\mathbf x_1,\mathbf x_2,\mathbf x_3,\mathbf x_4) \leftarrow \mathrm{Chunk}(\mathbf t, 4)$
    \STATE $\mathbf t_1 \leftarrow \mathrm{Concat}(-\mathbf x_2,\ \mathbf x_1,\ \mathbf x_4,\ -\mathbf x_3)$
    \STATE $\mathbf t_2 \leftarrow \mathrm{Concat}(\mathbf x_3,\ \mathbf x_4,\ \mathbf x_2,\ -\mathbf x_1)$
    \STATE $\mathbf t_3 \leftarrow \mathrm{Concat}(\mathbf x_4,\ -\mathbf x_3,\ -\mathbf x_1,\ -\mathbf x_2)$
  \STATE $\mathbf t \leftarrow \mathbf t \odot \cos
  \ +\ \mathbf t_1 \odot \frac{1}{3}\sin
  \ +\ \mathbf t_2 \odot \frac{2}{3}\sin
  \ +\ \mathbf t_3 \odot \frac{2}{3}\sin$
    \STATE \textbf{Output:} updated embedding $\mathbf t$
  \end{algorithmic}
\end{algorithm}

\subsection{Better Orthogonal Transformation $\mathbf Q$ Selection for HD-RoPE}
\label{sec.better_Orthogonal}

\paragraph{Conference matrices as structured orthogonal family.}

In the classical lineage of orthogonal matrix research, many constructions (such as Haar random orthogonality, Gram-Schmidt/QR generation, etc.) provide good isotropy, but often rely on randomness or numerical decomposition, leading to low reproducibility and additional overhead. We focus on a family of orthogonal matrices that are \textbf{\emph{scalable, numerically stable and reproducible, and provide sufficient cross-dimensional mixing}}.
Under these conditions, the available orthogonal families are limited: Hadamard/Schur type, discrete cosine/sine transforms, and sparse orthogonal parameterization based on givens rotations are all insufficient to meet the requirements. 
More details are in Appendix~\ref{app:prelim:structured_orthogonal}.

The \textbf{conference matrices} offer a compromise between theoretical properties and engineering friendliness: their discrete structures with elements of $\{0,\pm1\}$ make their implementations close to sparse addition and naturally exhibit strong cross-dimensional coupling. Therefore, we adopt the conference matrix as the primary source of construction for the orthogonal hybrid transform $\mathbf Q_N$ in HD-RoPE.
The N-order conference matrix $\mathbf C_N\in\{0,\pm 1\}^{N\times N}$ satisfies:
\begin{equation}
\mathrm{diag}(\mathbf C_N)=\mathbf 0,\quad \mathbf C_N\mathbf C_N^{\top}=(N-1)\mathbf I.
\end{equation}
The corresponding N-order orthogonal matrix can be obtained by:
\begin{equation}
\mathbf Q_N \triangleq \frac{1}{\sqrt{N-1}}\mathbf C_N,\quad
\mathbf Q_N^{\top}\mathbf Q_N=\mathbf I.
\end{equation}

This construction has two practical advantages: 
\begin{enumerate*}[label=(\roman*), itemjoin={{; }}, itemjoin*={{; and }}]
\item The element values of $\mathbf Q_N$ are \textbf{\emph{structured}} (only $0,\pm 1/\sqrt{N-1}$), making it easy to store and implement quickly (essentially a sign flip and sparse summation)
\item Given $N$ and the existence of a conference matrix of that order, $\mathbf Q_N$ can be directly generated by a deterministic construction algorithm, thus providing a \textbf{\emph{consistent depth-mixing coordinate transformation}} for HD-RoPE without the need for additional learning parameters.
\end{enumerate*}


\paragraph{Paley-I.}



Paley-I \cite{Paley1933OnOM} is a special conference matrices with good properties as $\bf{Q}$.
The special advantage of Paley-I lies in its balanced discrete structure. Its entries are restricted to $\{0,\pm 1\}$ before normalization, and each coordinate is mixed with all other coordinates through equal-magnitude coefficients after normalization. 
This yields a low-coordinate-bias basis: no single original channel dominates the transformed subspace, and the positional phase is distributed more uniformly across the $N$-dimensional block. 
Therefore, it provides a \textbf{\emph{deterministic, balanced, isotropic, and parameter-free basis}} in which higher-dimensional rotary phases can be applied \textbf{\emph{more uniformly and more sufficiently across query/key channels}}.  
Our HD-RoPE adopts Paley-I as $\bf{Q}$ in Eq. (\ref{eq.R_HD_RoPE}).

\paragraph{Dimension handling.} 

It is important to emphasize that the Paley-I construction for conference matrix does not exist for any dimension $N$; its existence is strongly constrained by combinatorial design and algebraic constraints. 
For example, the available dimension options below 32 include only: 2, 4, 8, and 32, but it is already sufficient to meet our main needs.
Details are in the Appendix~\ref{app:conference}.

\subsection{Our Higher-Dimensional RoPE}

\begin{table*}[ht]
\small
\begin{center}
\renewcommand\arraystretch{1.1}
\begin{tabular}{c|l|r|c|c|c|c|c|c}
\toprule
\multirow{2}{*}{\#Params} 
& \multicolumn{1}{c|}{\multirow{2}{*}{Rotation}}
& \multicolumn{1}{c|}{\multirow{2}{*}{$\theta$}}
& \multirow{2}{*}{HellaSwag} & \multirow{2}{*}{\ ARC-E\ } 
& \multirow{2}{*}{\ ARC-C\ } & \multirow{2}{*}{WinoGrande} 
& \multirow{2}{*}{\ BoolQ\ } & \multirow{2}{*}{\ AVG\ } \\
 &  &  &  &  &  &  &  &  \\
\midrule
\multirow{9.5}{*}{650M}
& \multirow{2}{*}{RoPE}
& $10^4$        & 26.69 & 48.87 & 26.58 & 51.29 & 52.09 & 41.10 \\
&  & $5 \times 10^5$ & 28.00 & 51.52 & 28.28 & 51.04 & 54.53 & \underline{42.68} \\
\cmidrule(lr){2-9}
& \multirow{2}{*}{3D-RPE}
& $10^4$        & 26.21 & 48.36 & 26.14 & 50.76 & 52.70 & 40.83 \\
&  & $5 \times 10^5$ & 27.41 & 50.91 & 27.63 & 50.52 & 54.10 & 42.11 \\
\cmidrule(lr){2-9}
& \multirow{2}{*}{FoPE}
& $10^4$        & 26.56 & 48.86 & 26.58 & 51.12 & 53.04 & 41.23 \\
&  & $5 \times 10^5$ & 27.86 & 51.51 & 28.15 & 51.23 & 54.64 & 42.67 \\
\cmidrule(lr){2-9}
& \multirow{2}{*}{HD-RoPE}
& $10^4$        & 27.39 & 50.53 & 28.25 & 51.39 & 54.76 & 42.46 \\
&  & $5 \times 10^5$ & 28.93 & 51.62 & 27.18 & 52.91 & 54.98 & \textbf{43.12} \\
\midrule

\multirow{9.5}{*}{1.3B}
& \multirow{2}{*}{RoPE}
& $10^4$        & 36.37 & 61.36 & 39.52 & 62.32 & 63.22 & 52.56 \\
&  & $5 \times 10^5$ & 37.20 & 63.21 & 40.62 & 63.11 & 62.73 & 53.37 \\
\cmidrule(lr){2-9}
& \multirow{2}{*}{3D-RPE}
& $10^4$        & 35.76 & 60.88 & 38.94 & 61.84 & 62.56 & 52.00 \\
&  & $5 \times 10^5$ & 36.54 & 62.63 & 39.97 & 62.54 & 62.10 & 52.76 \\
\cmidrule(lr){2-9}
& \multirow{2}{*}{FoPE}
& $10^4$        & 36.47 & 61.46 & 39.58 & 62.60 & 63.12 & 52.65 \\
&  & $5 \times 10^5$ & 37.16 & 63.02 & 40.67 & 63.27 & 62.78 & 53.38 \\
\cmidrule(lr){2-9}
& \multirow{2}{*}{HD-RoPE}
& $10^4$        & 37.68 & 62.57 & 40.95 & 63.58 & 63.01 & \underline{53.56} \\
&  & $5 \times 10^5$ & 41.43 & 63.45 & 42.54 & 64.59 & 66.47 & \textbf{55.70} \\
\bottomrule
\end{tabular}

\begin{minipage}{\linewidth}
\end{minipage}

\end{center}
\caption{Main results on general benchmarks with different model sizes and mainstream $\theta$ selections.}
\label{tab:exp:main} 
\end{table*}

\subsubsection{Rotation Matrix}

We attempt to construct a position-dependent rotation operator that (i) is orthogonal and norm-preserving,
(ii) preserves RoPE's relative-position structure, and (iii) induces stronger cross-channel coupling by rotating in mixed directions within an $N$-dimensional subspace.

Let the hidden dimension be $d$ and choose an even $N$.
We partition $\mathbb{R}^d$ into $B=d/N$ disjoint $N$-dimensional blocks and apply the same construction to each block.
Within an $N$-dimensional block, we first define a \emph{pairwise} (RoPE-isomorphic) rotation as:
\begin{equation}
\mathbf R_{\text{pair}}(p)
=
\mathrm{diag}\big(\mathrm{R}(\omega_1 p),\ldots,\mathrm{R}(\omega_{N/2} p)\big),
\end{equation}
\begin{equation}
\mathrm{R}(\alpha)=
\begin{pmatrix}
\cos\alpha & -\sin\alpha \\
\sin\alpha & \cos\alpha
\end{pmatrix}.
\end{equation}
To obtain a dense $N$-dimensional rotation with enhanced coupling, we perform an orthogonal change of basis as follows:
\begin{equation}
\mathbf R_{N}(p)\;\triangleq\;\mathbf Q_N^{\top}\,\mathbf R_{\text{pair}}(p)\,\mathbf Q_N, \quad \mathbf Q_N\in SO(N).\
\end{equation}
Therefore, attention scores continue to depend on the relative displacement $(s-p)$, while the rotation directions are no longer confined to fixed 2D coordinate planes, yielding richer directional couplings inside each $N$-dimensional subspace.

\subsubsection{A 4D Case}
\label{sec:method:a_4D_case}
Similar to RoPE, we first consider the simple case with dimension $d = 4$. The goal is to construct a function $g$ such that for any word vector $x_m, x_n\in\mathbb{R}^4$ and position $m, n\in\mathbb{Z}$. We have:
\begin{equation}
    \langle f_q(x_m,m), f_k(x_n,n)\rangle = g(x_m,x_n,m-n).
\end{equation}
Specifically, we further express $f\{q,k\}$ in the form of a multiplication matrix:
\begin{equation}
f_{\{q,k\}}(\boldsymbol{x}_m, n) = \mathbf{Q}^\top\mathbf{R}\mathbf{Q} \cdot \mathbf{W} \mathbf{x},
\end{equation}
where
\begin{equation}
    \mathbf{R} = \begin{pmatrix} \cos m\theta & -\sin m\theta & 0 & 0 \\ \sin m\theta & \cos m\theta  & 0 & 0 \\ 0&0& \cos m\theta & -\sin m\theta  \\ 0&0&\sin m\theta & \cos m\theta   \\ \end{pmatrix},
\end{equation}
\begin{equation}
    \mathbf{W} = \begin{pmatrix} W_{\{q,k\}}^{(11)}& \cdots & W_{\{q,k\}}^{(14)} \\ \vdots & \ddots &\vdots \\ W_{\{q,k\}}^{(41)}&\cdots & W_{\{q,k\}}^{(44)} \end{pmatrix},
\end{equation}
\begin{equation}
    \mathbf{x} = \begin{pmatrix} x_m^{(1)} & x_m^{(2)} & x_m^{(3)} & x_m^{(4)} \end{pmatrix}^\top.
\end{equation}
The orthogonal matrix $\mathbf Q_4$ is derived from the Paley-I conference matrix, noted as:
\begin{equation}
\mathbf C_{4} =
\begin{bmatrix}
0 & 1 & 1 & 1\\
1 & 0 & -1 & 1\\
1 & 1 & 0 & -1\\
1 & -1 & 1 & 0
\end{bmatrix},
\quad
\mathbf Q_4 = \frac{1}{\sqrt{3}}\mathbf C_4.
\end{equation}
The conjugated rotation $\mathbf Q_4^\top \mathbf R\mathbf Q_4$ preserves the norm and the relative-position closure with the desired characteristic stated in Section \ref{sec.better_Orthogonal}, replacing the original RoPE's pairwise phase interaction with a dense 4D phase interaction.

\subsubsection{General Form}

Let $\mathbf Q_4\in\mathbb{R}^{4\times 4}$ be an orthogonal matrix in a 4-dimensional subspace $(\mathbf Q_4^\top \mathbf Q_4=\mathbf I_4)$. Extending this to a block diagonal form in $d$ dimensions:
\begin{equation}
    \mathbf Q_{4}^{d}  \triangleq   \mathrm{diag}(\mathbf Q_4,\dots,\mathbf Q_4)\in\mathbb{R}^{d\times d}.
\end{equation}
The rotation matrix after the transformation corresponding to Section \ref{sec:method:a_4D_case} is:
\begin{equation}
\begin{aligned}
    \widetilde{\mathbf R}_{4,\Theta,m}^{d} &\triangleq (\mathbf Q_4^{d})^\top\,\mathbf R_{\Theta ,m}^{d}\,\mathbf Q_4^{d}\\
    &=  \mathrm{diag}(\mathbf R_{\theta_1,m}^{4},\dots,R_{\theta_{d/4},m}^{4})\in\mathbb{R}^{d\times d}. 
\end{aligned}
\end{equation}
We divide the d-dimensional space into d/4 subspaces and combine them using the linear property of the inner product, transforming $f\{q,k\}$ into:
\begin{equation}
\begin{aligned}
    f_{{q,k}}(\boldsymbol{x}_m, m) & = \widetilde{\mathbf R}_{4,\Theta,m}^{d}\,\mathbf W_{\{q,k\}}\,\boldsymbol{x}_m  \\& =  (\mathbf Q_4^{d})^\top\,\mathbf R_{4,\Theta,m}^{d}\,\mathbf Q_4^{d}\,\mathbf W_{\{q,k\}}\,\boldsymbol{x}_m.
\end{aligned}
\end{equation}


Similarly, we can extend $\mathbf Q_N$ to $d$ dimensions by block-diagonal replication as $\mathbf Q_N^{d}$.
The resulting $d$-D rotation takes form as:
\begin{equation}
\begin{aligned}
    \widetilde{\mathbf R}_{N,\Theta,m}^{d} &\triangleq (\mathbf Q_N^{d})^\top\,\mathbf R_{N,\Theta ,m}^{d}\,\mathbf Q_N^{d}\\
    &=  \mathrm{diag}(\mathbf R_{\theta_1,m}^{N},\dots,R_{\theta_{d/N},m}^{N})\in\mathbb{R}^{d\times d}. 
\end{aligned}
\end{equation}


\subsection{Implementation and Efficiency}\label{sec:method:implement}


Since $\mathbf Q$ is constant and orthogonal, each entry of $\widetilde{\mathbf R}(m)$ remains a fixed linear combination of $\sin(\cdot)$ and $\cos(\cdot)$ terms, while the use of structured conference matrices yields a \emph{dense} mixed rotation (with very few zero entries).

This structure enables the same engineering optimization as RoPE: we precompute the required trigonometric terms and apply the rotation through linear operations that emulate the matrix product, substantially reducing the overhead of HD-RoPE. The full procedure is summarized in Algorithm \ref{alg:hdrope-4d-paleyi-freqs}.
We provide a more in-depth analysis of the effectiveness of HD-RoPE in the appendix. 

%




\section{Experiment}


\subsection{Experiment Setup}\label{sec:experiment_setup}

\begin{table*}[!ht]
\small
\begin{center}
\renewcommand\arraystretch{1.1}
\begin{tabular}{c|c|r|c|c|c|c|c}
\toprule
\multirow{2}{*}{\#Params} & \multirow{2}{*}{Rotation} & 
\multicolumn{1}{c|}{\multirow{2}{*}{$\theta$}}
& \multirow{2}{*}{Passkey-Retrieval} 
& \multirow{2}{*}{Stack-Select} 
& \multirow{2}{*}{TextSort} 
& \multirow{2}{*}{LongBench-v2} 
& \multirow{2}{*}{AVG} \\
 &  &  &  &  &  &  &  \\
\midrule

\multirow{4.5}{*}{650M}
& \multirow{2}{*}{2D}
& $10^4$        & 42.49 & 39.43 & 29.58 & 32.85 & 36.09 \\
&  & $5 \times 10^5$ & 41.87 & 38.96 & 30.94 & 31.29 & 35.77 \\
\cmidrule(lr){2-8}
& \multirow{2}{*}{4D}
& $10^4$        & 41.37 & 38.49 & 30.42 & 34.13 & 36.10 \\
&  & $5 \times 10^5$ & 40.04 & 41.13 & 32.07 & 35.72 & \textbf{37.24} \\
\midrule

\multirow{4.5}{*}{1.3B}
& \multirow{2}{*}{2D}
& $10^4$        & 58.84 & 49.79 & 46.87 & 43.66 & 49.79 \\
&  & $5 \times 10^5$ & 60.50 & 51.29 & 48.31 & 45.22 & 51.33 \\
\cmidrule(lr){2-8}
& \multirow{2}{*}{4D}
& $10^4$        & 60.53 & 50.67 & 47.76 & 44.94 & 50.97 \\
&  & $5 \times 10^5$ & 62.10 & 52.12 & 49.06 & 46.52 & \textbf{52.45} \\
\bottomrule
\end{tabular}
\end{center}
\caption{Evaluation results on long-context benchmarks with different model sizes and $\theta$ selections.}
\label{tab:exp:long_eval}
\end{table*}

\noindent
\textbf{Model setting and training configuration.}
We trained from scratch on a \textsc{Llama-3} \cite{grattafiori2024llama} style decoder-only Transformer architecture, training two models of different sizes: 650M and 1.3B. All models in Table \ref{tab:exp:main} were trained with 50B tokens.
All comparative experiment settings maintained strict consistency, and the details are presented in Appendix \ref{app:model_config}.

\noindent
\textbf{Pre-training dataset.}
The pre-training data used the open-source SlimPajama \cite{shen2023slimpajama} corpus. We used the Meta-rater \cite{zhuang2025meta} quality scoring to screen SlimPajama, and selected the Readability-30B, Professionalism-30B, and Reasoning-30B subsets for deduplication and merging to form the final training set.

\noindent
\textbf{Baselines and evaluations.}
The major baseline is the standard 2D RoPE \cite{su2024roformer}. We systematically scanned the RoPE base $\theta$ in the range of $[1\mathrm{k}, 5\mathrm{M}]$. In the main comparison, we report representative and best-performing settings, including $\theta=10\mathrm{k}$ and $\theta=500\mathrm{k}$, which are widely used in existing LLMs.
We also compare with 3D-RPE \cite{ma20253d} and FoPE \cite{hua2024fourier}.
We evaluate models on five short-context benchmarks and three long-context benchmarks. More details are in Appendix \ref{app.baseline}.


\subsection{Main Results on General Benchmarks}

\begin{table}[t]
    \small
  \begin{center}
        \begin{tabular}{c|c|r|c}
          \toprule
          \multirow{2}{*}{Rotation}& \multirow{2}{*}{\begin{tabular}[c]{@{}c@{}}Orthogonal\\ Matrix ($\mathbf Q$)\end{tabular}}  & \multicolumn{1}{c|}{\multirow{2}{*}{$\theta$}} &\multicolumn{1}{c}{\multirow{2}{*}{Short-Text AVG}}      \\
          &&& \\ 
          \midrule 
      \multirow{2}{*}{2D} & \multirow{2}{*}{Vanilla} & $10^4$ & 51.70   \\
      &&{${5 \times 10^5}$}&52.49\\
          \cmidrule(lr){1-4}
      \multirow{7}{*}{4D}  & \multirow{2}{*}{\begin{tabular}[c]{@{}c@{}} Identity\\($\mathbf Q=\mathbf I$)\end{tabular}} &$10^4$ &51.04 \\
      &&{${5 \times 10^5}$}&50.91 \\
      \cmidrule(lr){2-4}
       & \multirow{2}{*}{\begin{tabular}[c]{@{}c@{}} Random\end{tabular}} & $10^4$& 35.12 \\
      &&{${5 \times 10^5}$}&37.92 \\
          \cmidrule(lr){2-4}
         & \multirow{2}{*}{\begin{tabular}[c]{@{}c@{}} \textbf{Conference} \\\textbf{(Paley-I)}\end{tabular}} & $10^4$&51.95  \\
      && {${5 \times 10^5}$ }&53.74 \\
          \bottomrule
        \end{tabular}
   \begin{minipage}{\linewidth}
\end{minipage}
  \end{center}
  \caption{Ablation Study of Rotation for HD-RoPE.}
  \label{tab:exp:ablation}
\end{table}

\begin{figure}[!t]
  \begin{center}
    \centerline{\includegraphics[width=0.95\columnwidth]{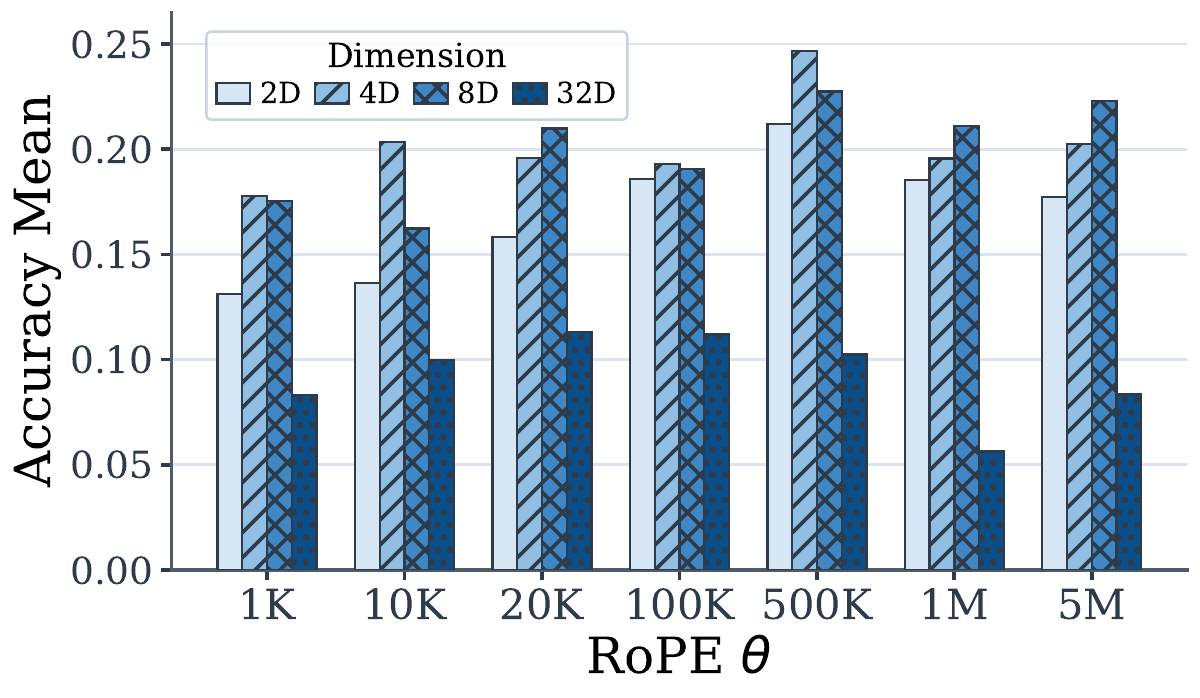}}
    \caption{Rotation dimension--$\theta$ sweep. The mean accuracy is reported for 2D, 4D, 8D, and 32D rotations under different RoPE base values.}
    \label{fig:exp:theta-sweep}
  \end{center}
\end{figure}

\noindent
\textbf{HD-RoPE outperforms RoPE.}
Table~\ref{tab:exp:main} shows the main results on the short-context benchmarks. 
HD-RoPE outperformed 2D RoPE on the vast majority of specific tasks and average results. At the same time, HD-RoPE also surpasses other baselines, demonstrating consistency across different tasks, model sizes, and typical RoPE $\theta$.

\noindent
\textbf{Stable gains with various $\theta$.}
This performance improvement is not limited solely to the two specific $\theta$ values mentioned above.
Under comparable $\theta$ settings, our 4D HD-RoPE variant consistently improves the average score over 2D RoPE. Details are shown in Section \ref{sec:exp:thetas-NDs}.


\noindent
\textbf{Efficiency and usability.}
As in Section \ref{sec:method:implement}, our implementation references the engineering optimizations of RoPE, performing similar matrix multiplications for linear complexity.
HD-RoPE can be applied plug-and-play to various models. 

\subsection{Long-Context Evaluations}

Table~\ref{tab:exp:long_eval} reports the long-context evaluation results. The results confirm that HD-RoPE improves long-context performance in both model scales, suggesting that our high-dimensional rotation is beneficial for tasks requiring long-range dependency modeling and long-context reasoning.

To further evaluate HD-RoPE's long-context model ability, we implement our model and baseline with popular long-context extrapolation methods, i.e., YaRN \cite{peng2023yarn} and LongRoPE \cite{ding2024longrope}. The results in Table \ref{tab:exp:long_eval_extrapolation} clearly indicate that HD-RoPE achieves better overall performance compared to the original 2D RoPE on long-context benchmarks, implying better potential of HD-RoPE in practice.


\begin{table*}[!ht]
\small
\begin{center}
\renewcommand\arraystretch{1.1}
\begin{tabular}{c|c|l|c|c|c|c|c}
\toprule
\multirow{2}{*}{\#Params} 
& \multirow{2}{*}{Rotation} 
& \multicolumn{1}{c|}{\multirow{2}{*}{Method}}
& \multirow{2}{*}{Passkey-Retrieval} 
& \multirow{2}{*}{Stack-Select} 
& \multirow{2}{*}{TextSort} 
& \multirow{2}{*}{LongBench-v2} 
& \multirow{2}{*}{AVG} \\
 &  &  &  &  &  &  &  \\
\midrule

\multirow{4}{*}{650M}
& \multirow{2}{*}{2D}
& YaRN     & 39.72 & 37.84 & 29.91 & 30.44 & 34.48 \\
& 
& LongRoPE & 40.58 & 38.21 & 30.15 & 30.88 & 34.96 \\
\cmidrule(lr){2-8}
& \multirow{2}{*}{4D}
& YaRN     & 39.25 & 39.88 & 31.22 & 34.21 & 36.14 \\
& 
& LongRoPE & 39.61 & 40.42 & 31.58 & 34.76 & \textbf{36.59} \\
\midrule

\multirow{4}{*}{1.3B}
& \multirow{2}{*}{2D}
& YaRN     & 58.73 & 50.14 & 47.23 & 44.16 & 50.07 \\
& 
& LongRoPE & 59.31 & 50.63 & 47.81 & 44.58 & 50.58 \\
\cmidrule(lr){2-8}
& \multirow{2}{*}{4D}
& YaRN     & 60.24 & 50.95 & 48.06 & 45.14 & 51.10 \\
& 
& LongRoPE & 60.91 & 51.38 & 48.54 & 45.61 & \textbf{51.61} \\
\bottomrule
\end{tabular}
\end{center}
\caption{Evaluation results with long-context extrapolation methods (YaRN and LongRoPE), where $\theta = 5 \times 10^5$.}
\label{tab:exp:long_eval_extrapolation}
\end{table*}

\begin{table*}[!ht]
\small
\begin{center}
\renewcommand\arraystretch{1.1}
\begin{tabular}{c|c|c|c|c|c|c}
\toprule
\multirow{2}{*}{\#Params} & \multirow{2}{*}{Rotation} 
& \multirow{2}{*}{Passkey-Retrieval} 
& \multirow{2}{*}{Stack-Select} 
& \multirow{2}{*}{TextSort} 
& \multirow{2}{*}{LongBench-v2} 
& \multirow{2}{*}{AVG} \\
 &  &  &  &  &  &  \\
\midrule
\multirow{2}{*}{1.3B}
& 2D
&  61.27 & 54.79 & 50.13 & 45.53 & 52.93 \\
& 4D
&  64.12 & 55.29 & 50.51 & 46.91 & \textbf{54.21} \\
\bottomrule
\end{tabular}
\end{center}
\caption{Evaluation results for the 1.3B model with long-context continue pre-training, $\theta = 5 \times 10^5$.}
\label{tab:exp:long_eval_cont_pre_train}
\end{table*}

\subsection{Ablation Study}


\begin{figure*}[!htp]
  \centering
  \setlength{\tabcolsep}{2pt}
  \renewcommand{\arraystretch}{8.0}
  \begin{tabular}{cccc}
    \begin{subfigure}[t]{0.23\textwidth}
      \centering
      \includegraphics[width=\linewidth]{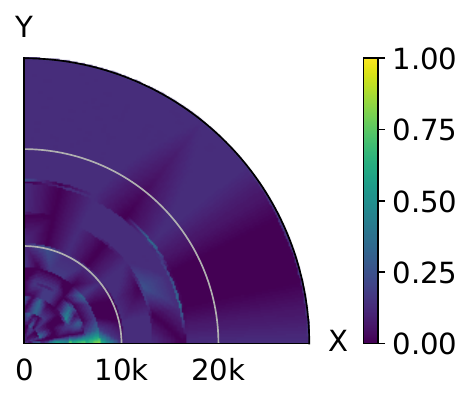}
      \caption{2D RoPE.}
      \label{fig:compare-a}
    \end{subfigure} &
    \begin{subfigure}[t]{0.23\textwidth}
      \centering
      \includegraphics[width=\linewidth]{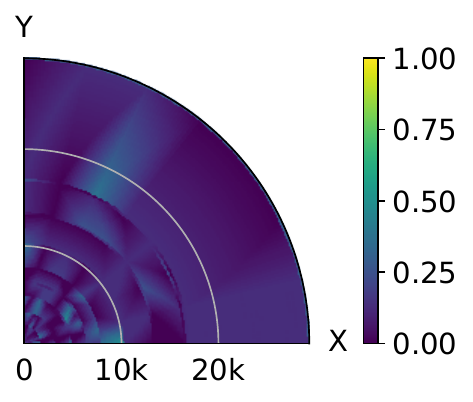}
      \caption{4D HD-RoPE.}
      \label{fig:compare-b}
    \end{subfigure} &
    \begin{subfigure}[t]{0.23\textwidth}
      \centering
      \includegraphics[width=\linewidth]{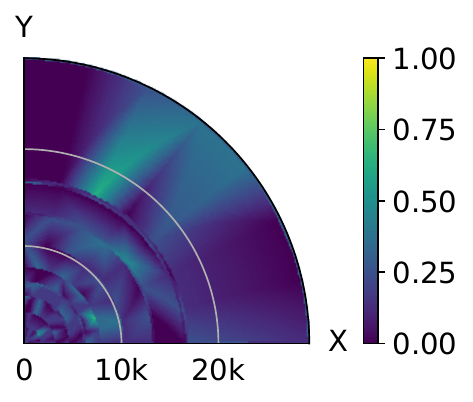}
      \caption{8D HD-RoPE.}
      \label{fig:compare-c}
    \end{subfigure} &
    \begin{subfigure}[t]{0.23\textwidth}
      \centering
      \includegraphics[width=\linewidth]{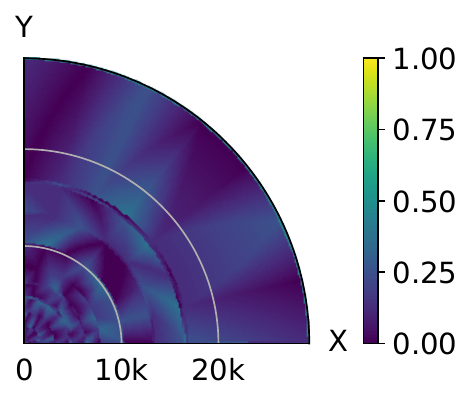}
      \caption{32D HD-RoPE.}
      \label{fig:compare-d}
    \end{subfigure}
  \end{tabular}
  \caption{Passkey-position profile of Passkey Retrieval accuracy under fixed RoPE base $\theta=1\mathrm{k}$, visualized as polar fan heatmaps.}
  \label{fig:exp:passkey-position}
\end{figure*}

To isolate the contribution of our HD-RoPE, we conduct an ablation study during the transition from standard 2D RoPE to 4D HD-RoPE, while keeping all other training and evaluation configurations identical. All models in the ablation study were trained with 30B tokens. Specifically, we compare the following three settings: 
\begin{enumerate*}[label=(\roman*), itemjoin={{; }}, itemjoin*={{; and }}]
\item \emph{\textbf{4D grouping with pairwise RoPE (no mixing).}}
We still apply the standard 2D pairwise RoPE rotation, but assign frequencies and organize channels in 4D groups. This corresponds to a degenerate 4D formulation with an identity orthogonal transform, i.e., $\mathbf Q=\mathbf I$
\item \emph{\textbf{4D + random orthogonal mixing.}}
$\mathbf Q$ is obtained by random orthogonalization to test whether orthogonality alone is sufficient to yield improvements
\item \emph{\textbf{4D + conference mixing (Paley-I; our final version).}}

\end{enumerate*}

The results are reported in Table \ref{tab:exp:ablation}. Transitioning from standard 2D RoPE to \emph{4D grouping with pairwise RoPE} yields only limited gains, indicating that merely modifying the frequency indexing without introducing cross-dimensional mixing is insufficient to obtain substantial improvements. Moreover, \emph{4D + random orthogonal mixing} does not consistently improve over baseline and can even degrade performance, suggesting that \emph{orthogonality alone} is not the key factor: unstructured mixing may disrupt the alignment between frequency assignment and channel geometry. The gains of HD-RoPE primarily arise from \emph{structured, deterministic, and strongly mixing} orthogonal conjugation, rather than regrouping or unstructured orthogonal transformations.



\subsection{Effect of Different RoPE Base $\theta$ Across Various Rotation Dimensions}\label{sec:exp:thetas-NDs}

We further study how the RoPE base $\theta$ interacts with the rotation dimension. 
Figure~\ref{fig:exp:theta-sweep} aggregates the results across different $\theta$ values. The 4D setting achieves the best overall performance, with 8D being close but less stable across $\theta$ sweep. 
In contrast, 32D performs substantially worse, suggesting that simply increasing the rotation dimension is not sufficient. Excessively high-dimensional rotation may over-mix the channel structure, or make the effective frequency allocation less suitable for the model. 
These results support 4D HD-RoPE as an efficient sweet spot: it introduces sufficient cross-channel coupling while avoiding the degradation observed in overly large rotation groups.

Moreover, we also observe that 4D HD-RoPE consistently outperforms 2D RoPE with different $\theta$, reconfirming the effectiveness of our method.

\begin{figure}[!t]
  \begin{center}
    \centerline{\includegraphics[width=0.75\columnwidth]{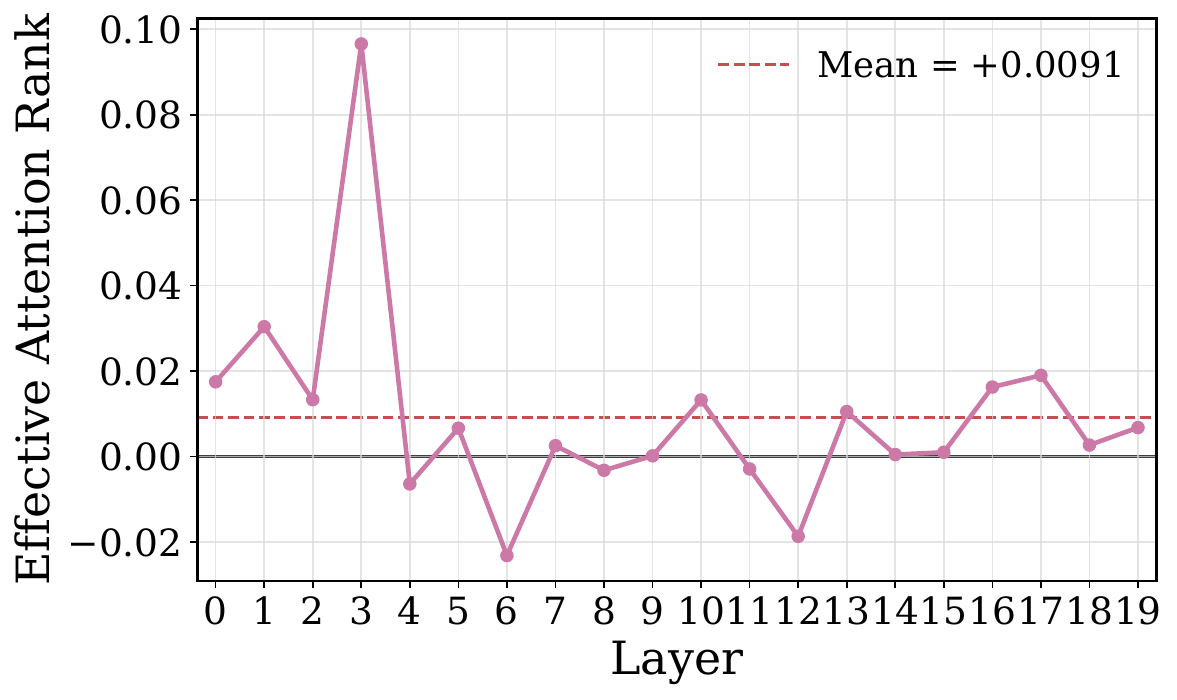}}
    \caption{Layer-wise effective attention-rank difference between HD-RoPE and 2D RoPE. }
    \label{fig:exp:effective-rank}
  \end{center}
\end{figure}

\subsection{Effect of HD-RoPE with Long-Context Continue Pre-training}



Based on the model used in our main experiments, we conducted long-context continued pre-training to verify the effectiveness of HD-RoPE, utilizing a sequence length of 32,768 and a total of 13.4B training tokens. As shown in Table \ref{tab:exp:long_eval_cont_pre_train}, HD-RoPE maintained its performance advantage.

\subsection{Mechanistic Evidence for HD-RoPE}

To better understand why HD-RoPE improves long-context behavior, we analyze both the internal attention structure and the external retrieval pattern. 
Figure~\ref{fig:exp:passkey-position} provides a complementary task-level view. Under fixed RoPE base $\theta=1\mathrm{k}$, the radial coordinate represents sequence length and the polar angle denotes the relative position of the answer. Compared with 2D RoPE, higher-dimensional rotations expand the high-accuracy region toward the middle part of the sequence, where retrieval is typically more difficult. 
Figure~\ref{fig:exp:effective-rank} shows the layer-wise effective attention-rank difference between 4D HD-RoPE and 2D RoPE. The mean difference is positive, suggesting that HD-RoPE tends to produce less collapsed and more diverse attention patterns. 




\section{Related Works}

Position modeling in Transformers is generally categorized into three types: absolute position encoding, relative position encoding, and explicit biases. Classical Transformers employ sinusoidal absolute method \cite{vaswani2017attention}; subsequent work shifted towards direct forms, such as introducing relative distance representations \cite{shaw2018self} or sharing bias parameters across different distance \cite{raffel2020exploring}. ALiBi \cite{press2021train} demonstrates length extrapolation capabilities through negative biases.

Rotary Position Embedding (RoPE; \citealp{su2024roformer}) makes the attention inner product explicitly dependent on the relative displacement.  Recent works is mainly \emph{recalibrating the position index or frequency}. Position Interpolation (PI) \cite{chen2023extending} reduces the phase curse by linearly compressing the position index. YaRN \cite{peng2023yarn} further proposes a more efficient RoPE windowing scheme. LongRoPE combines interpolation search, extending the available context\cite{ding2024longrope}. Other works have analyzed the extrapolation patterns of RoPE from periodicity and scale \cite{liu2023scaling}, 
or provided a explanation \cite{zhong2025understanding} of the mechanism of RoPE extension methods.

Recent works rethinks positional encoding based on representation morphology. FoPE \cite{hua2024fourier} interprets RoPE as an implicit discrete Fourier transform. PoPE \cite{gopalakrishnan2025decoupling} explicitly separates content and position in polar coordinates. DroPE \cite{gelberg2025extending} removes positional embeddings after pre-training.

\section{Conclusion}
We propose HD-RoPE as a novel position embedding, focusing on the rotational degrees of freedom itself. By extending the two-dimensional rotation of RoPE to higher-dimensional subspaces and using structured orthogonal transformations to achieve cross-dimensional mixing, the expressiveness of position representation is greatly improved while preserving the relative-position properties.

\section*{Limitations}
HD-RoPE focuses on enhancing RoPE by replacing isolated 2D rotations with higher-dimensional rotation subspaces. This design increases the channel capacity and mixing density of each positional phase, but it also reduces the number of independent frequency groups under a fixed hidden dimension. Therefore, HD-RoPE introduces a trade-off between fine-grained frequency diversity and richer within-frequency channel interaction. In addition, our current implementation uses fixed Paley-I conference matrices, whose construction only exists for specific dimensions and may not be universally optimal. Future work may investigate more flexible orthogonal bases, adaptive mixing strategies, and a deeper theoretical understanding of the optimal rotation dimension.




\bibliography{custom}

\appendix

\section{Appendix}
\label{sec:appendix}

\subsection{Supplement to Related Works}
Beyond the windowing approach based on frequency recalibration, recent research has begun to reconstruct the feasible domain and generalizability of RoPE from an algebraic perspective. LieRE \cite{ostmeier2024liere} extends rotation encoding to a more general $n$-dimensional input domain and validates the effectiveness of relative position rotation in non-textual modalities in 2D/3D vision tasks. \cite{liu2025rethinking} further presents a unified mathematical blueprint for RoPE, emphasizing constraints such as relativity on high-dimensional constructions and pointing out that cross-dimensional interactions can be characterized by learning/selecting appropriate orthogonal basis transformations. Complementary to this, \cite{yu2025comrope} starts from the RoPE equations, parameterizing rotations into trainable angle matrices and emphasizing the crucial role of commutativity in maintaining translation consistency and scalability. These works collectively demonstrate that the generalization of RoPE is not merely an engineering technique, but a problem strongly constrained by algebraic structures. The HD-RoPE in this paper employs deterministic structured orthogonal hybridization to achieve high-dimensional deep rotation, enhancing cross-dimensional coupling while maintaining the relative displacement closure.

Regarding the specific forms of high-dimensional rotation, 3D-RPE \cite{ma20253d} elevates rotation from a two-dimensional plane to a three-dimensional spherical parameterization, improving long-range attenuation control and position resolution. Meanwhile, GeoPE \cite{yao2025geope} roposes a geometrically coupled rotation based on quaternions and Lie algebra means for structured tensors (2D/3D), emphasizing cross-axis joint geometric consistency. In terms of vision and multimodal applications, \cite{heo2024rotary} systematically evaluates the implementation details and resolution extrapolation performance of RoPE in ViT. Qwen2-VL \cite{wang2024qwen2} introduces M-RoPE to unify the fusion of position signals from text, images, and videos, supporting dynamic resolution visual tokenization. Work \cite{chiang2025rotary} indicates that large-angle rotations of RoPEs in long-distance scenarios may lead to a decrease in the utilization of some dimensions, suggesting that the structure and hybridization of the rotated subspace itself affects long-range retrieval capabilities. These results further support the motivation of this paper: to improve available dimensions and cross-dimensional information interaction through higher-dimensional rotated spaces and structured orthogonal hybridization while maintaining the relative constraints of RoPEs.

\subsection{Detailed Baselines and evaluations.}\label{app.baseline}

The major baseline is the standard 2D RoPE \cite{su2024roformer}. We systematically scanned the RoPE base $\theta$ in the range of $[1\mathrm{k}, 5\mathrm{M}]$. In the main comparison, we report representative and best-performing settings, including $\theta=10\mathrm{k}$ and $\theta=500\mathrm{k}$, which are widely used in existing LLMs.
We also compare with 3D-RPE \cite{ma20253d} and FoPE \cite{hua2024fourier}.

We evaluate models on four short-context benchmarks and three long-context benchmarks. First, for the main short-context evaluation, we report standard benchmark results: (1) Hellaswag \cite{zellers2019hellaswag}; (2) AI2 Reasoning Challenge \citep{clark2018think}; (3) WinoGrande \cite{sakaguchi2021winogrande}; (4) BoolQ \cite{clark2019boolq}. Second, for long-context evaluation: (1) Passkey-Retrieval, measuring precise retrieval capabilities in long sequences, following the specifications of the LongRoPE \cite{ding2024longrope}; (2) Ada-LEval \cite{wang2024ada}, covering task types more sensitive to long dependencies, including two subtasks: stackselect and textsort; (3) LongBench-v2 \cite{bai2025longbench}, emphasizing long context understanding and reasoning in real-world multi-task scenarios. Note that the scores of LongBench-v2 are computed from logits, i.e. \textsc{LOGIT SCORES}, not accuracy. Best in bold; second best underlined.

\subsection{Model and Training Settings}\label{app:model_config}

This appendix provides the detailed model and training configurations used in our experiments. We will open-source the training and data framework in a future version.

\subsubsection{Model Architecture}

We train two Llama-style decoder-only Transformer models from scratch, with sizes of 650M and 1.3B parameters. The architectural configurations are summarized in Table~\ref{tab:app:model_arch}. Both models use RMSNorm, SwiGLU feed-forward networks, and Gaussian weight initialization with standard deviation 0.02.

\begin{table}[h]
\centering
\small
\renewcommand{\arraystretch}{1.1}
\begin{tabular}{c|c|c}
\toprule
Configuration & 650M & 1.3B \\
\midrule
\#Params & 650M & 1.3B \\
Hidden Dim. & 1536 & 2048 \\
Layers & 16 & 20 \\
Heads & 24 & 32 \\
KV Heads & 6 & 8 \\
Head Dim. & 64 & 64 \\
Normalization & RMSNorm & RMSNorm \\
FFN Type & SwiGLU & SwiGLU \\
Weight Initialization & $\mathcal{N}(0, 0.02)$ & $\mathcal{N}(0, 0.02)$ \\
\bottomrule
\end{tabular}
\caption{Model architecture configurations.}
\label{tab:app:model_arch}
\end{table}

\subsubsection{Training Configuration}

The training data construction follows Section~\ref{sec:experiment_setup}. All models are trained with the same training budget and optimization configuration for fair comparison. We use a global batch size of 64 and train each model on approximately 50B tokens. Training is conducted with bf16 mixed precision and DeepSpeed ZeRO-2.

For optimization, we use AdamW with the WarmupDecayLR scheduler implemented in DeepSpeed. The learning rate is warmed up from $1\times10^{-6}$ to $1\times10^{-4}$, followed by decay according to the scheduler. All model variants share the same optimizer, scheduler, precision, batch size, and training-token budget.

All models and components are configured with a unified random seed of 42 to ensure fair comparison. Specifically, the seeds for \texttt{random}, \texttt{numpy}, and \texttt{torch} are all fixed to 42.

\subsubsection{Attention Implementation}

The attention computation uses PyTorch SDPA. RoPE or HD-RoPE is applied before the query and key tensors enter the attention kernel, and is therefore decoupled from the kernel implementation:
\begin{equation}
\begin{aligned}
    \mathbf{Q}, \mathbf{K} &= \mathrm{ApplyRotaryEmb}(\mathbf{Q}, \mathbf{K}),\\ 
    \mathbf{O} &= \mathrm{SDPA}(\mathbf{Q}, \mathbf{K}, \mathbf{V}).
\end{aligned}
\end{equation}
This implementation allows HD-RoPE to be integrated into the standard attention pipeline without modifying the SDPA kernel.

\subsubsection{Computational Overhead}

We report the FLOPs overhead using the 1.3B model with sequence length 8192 and micro-batch size 1 as an example. The total model FLOPs for a single forward pass, excluding RoPE, is approximately 21.7T. Compared with the main model computation, the additional FLOPs introduced by RoPE and HD-RoPE are negligible.

\begin{table}[h]
\centering
\small
\renewcommand{\arraystretch}{1.1}
\begin{tabular}{c|c|c}
\toprule
Setting & Total FLOPs & Proportion vs. Model \\
\midrule
2D RoPE & 1.26G & 0.0058\% \\
4D HD-RoPE & 2.94G & 0.0135\% \\
\bottomrule
\end{tabular}
\caption{Additional FLOPs introduced by positional rotation.}
\label{tab:app:flops}
\end{table}

We further measure GPU memory usage under 8 GPUs, sequence length 8192, and DeepSpeed ZeRO-2. As shown in Table~\ref{tab:app:memory}, HD-RoPE introduces almost no additional memory overhead compared with standard 2D RoPE.

\begin{table}[h]
\centering
\small
\renewcommand{\arraystretch}{1.1}
\begin{tabular}{c|c|c}
\toprule
Setting & 650M / GB & 1.3B / GB \\
\midrule
w/o PE & 19.69 & 29.83 \\
2D RoPE & 19.71 & 29.87 \\
4D HD-RoPE & 19.72 & 29.87 \\
2D--4D Interleave & 19.71 & 29.87 \\
\bottomrule
\end{tabular}
\caption{GPU memory usage under sequence length 8192 and ZeRO-2.}
\label{tab:app:memory}
\end{table}

For runtime efficiency, we report the relative change in step time and throughput for the 1.3B model. HD-RoPE only introduces minor latency overhead, while maintaining practical training efficiency.

\begin{table}[h]
\centering
\small
\renewcommand{\arraystretch}{1.1}
\begin{tabular}{c|c|c}
\toprule
Machine Setting & Step Time & Throughput \\
\midrule
1 Machine / 8 GPUs & +2.58\% & -2.54\% \\
4 Machines / 32 GPUs & +1.23\% & -1.10\% \\
\bottomrule
\end{tabular}
\caption{Runtime overhead of HD-RoPE on the 1.3B model.}
\label{tab:app:latency}
\end{table}

Overall, HD-RoPE introduces negligible FLOPs and memory overhead, and only minor runtime latency, making it practical for standard large-scale pre-training settings.

\subsection{FC Attention Magnitude of RoPE and HD-RoPE}

We also attempted to analyze how HD-RoPE influences attention. Shown below is the amplitude plot following Fourier decomposition.

\begin{figure}[!t]
  \begin{center}
    \centerline{\includegraphics[width=0.95\columnwidth]{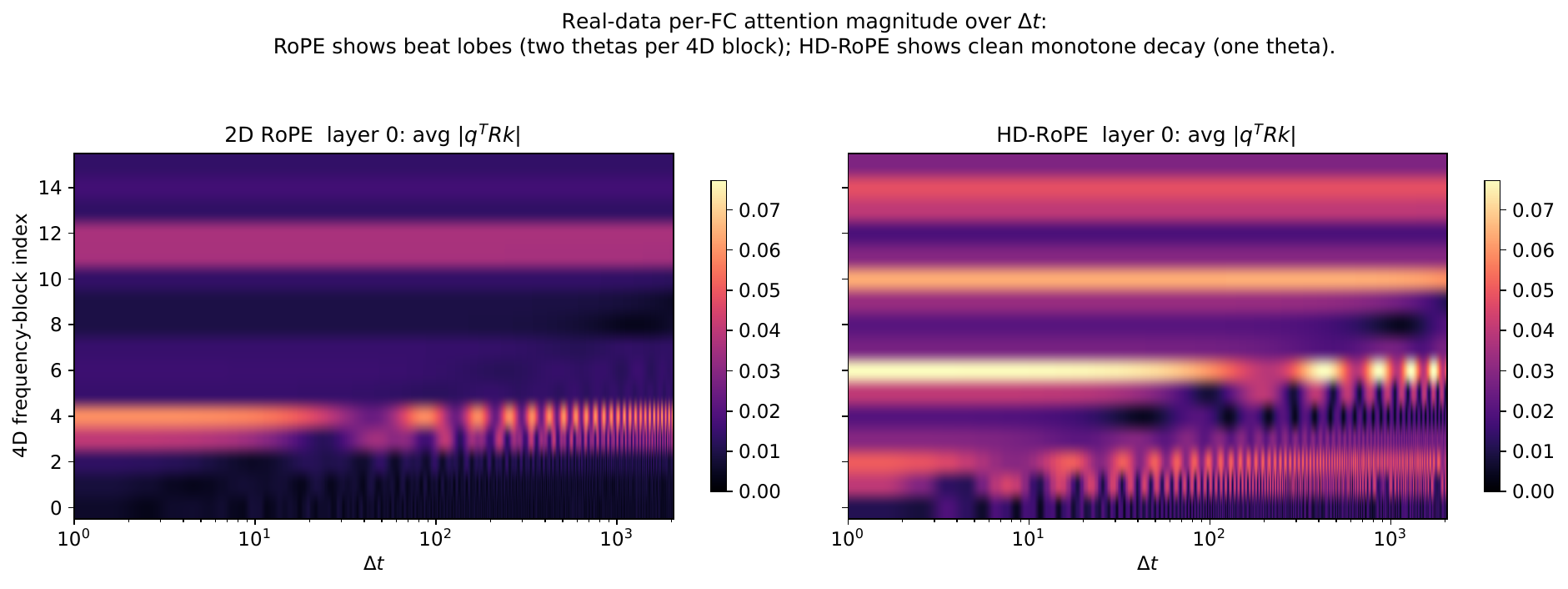}}
    \caption{FC Attention Magnitude of RoPE and HD-RoPE, layer 0. }
    \label{fig:magl0}
  \end{center}
\end{figure}

\begin{figure}[!t]
  \begin{center}
    \centerline{\includegraphics[width=0.95\columnwidth]{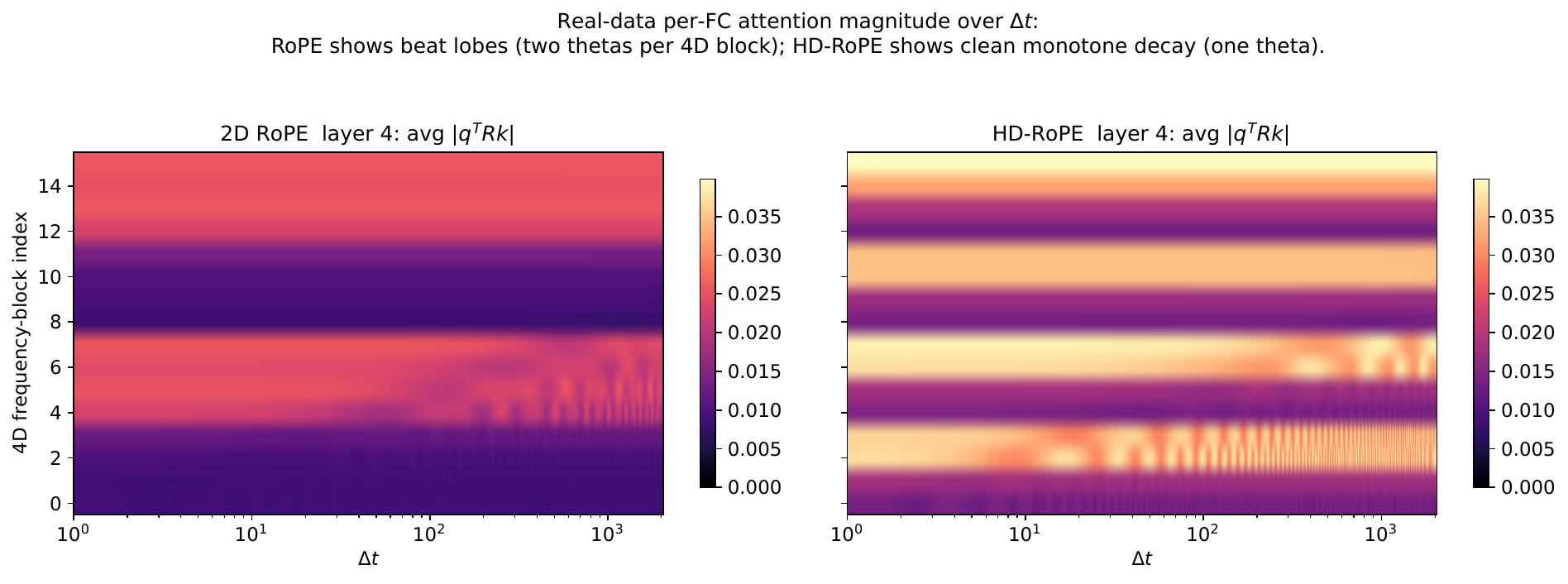}}
    \caption{FC Attention Magnitude of RoPE and HD-RoPE, layer 4. }
    \label{fig:magl0}
  \end{center}
\end{figure}

\begin{figure}[!t]
  \begin{center}
    \centerline{\includegraphics[width=0.95\columnwidth]{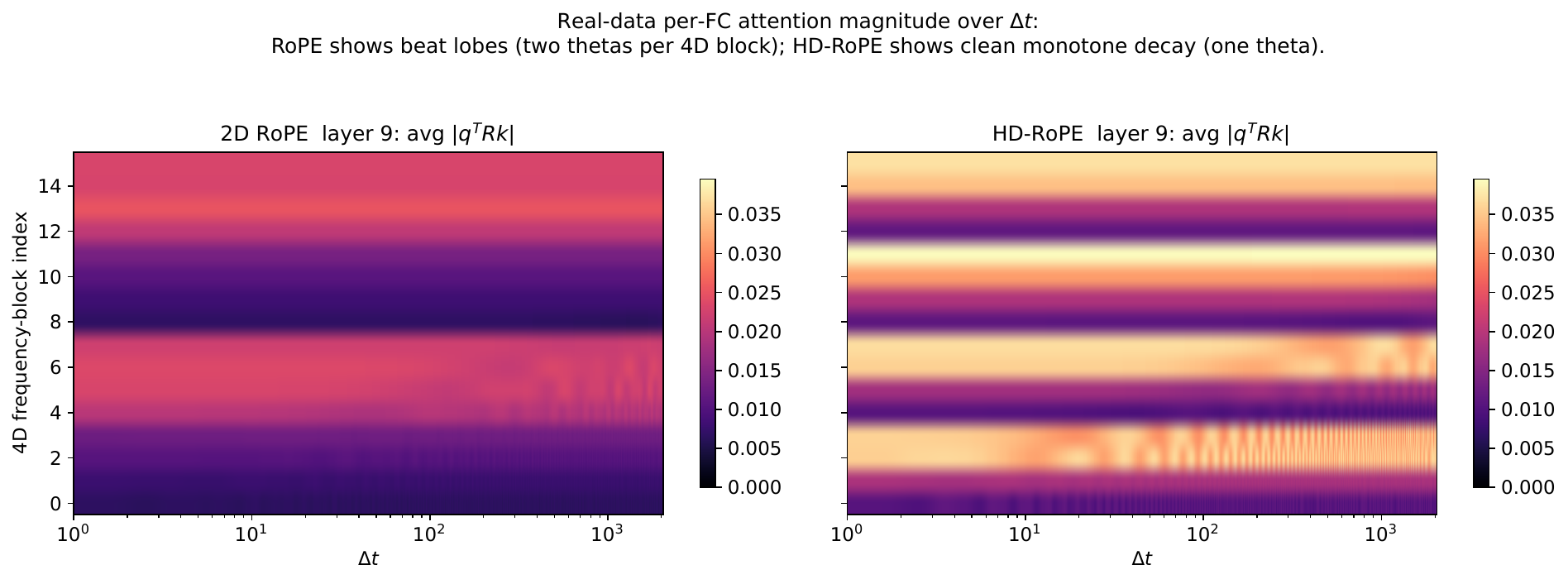}}
    \caption{FC Attention Magnitude of RoPE and HD-RoPE, layer 9. }
    \label{fig:magl0}
  \end{center}
\end{figure}

\begin{figure}[!t]
  \begin{center}
    \centerline{\includegraphics[width=0.95\columnwidth]{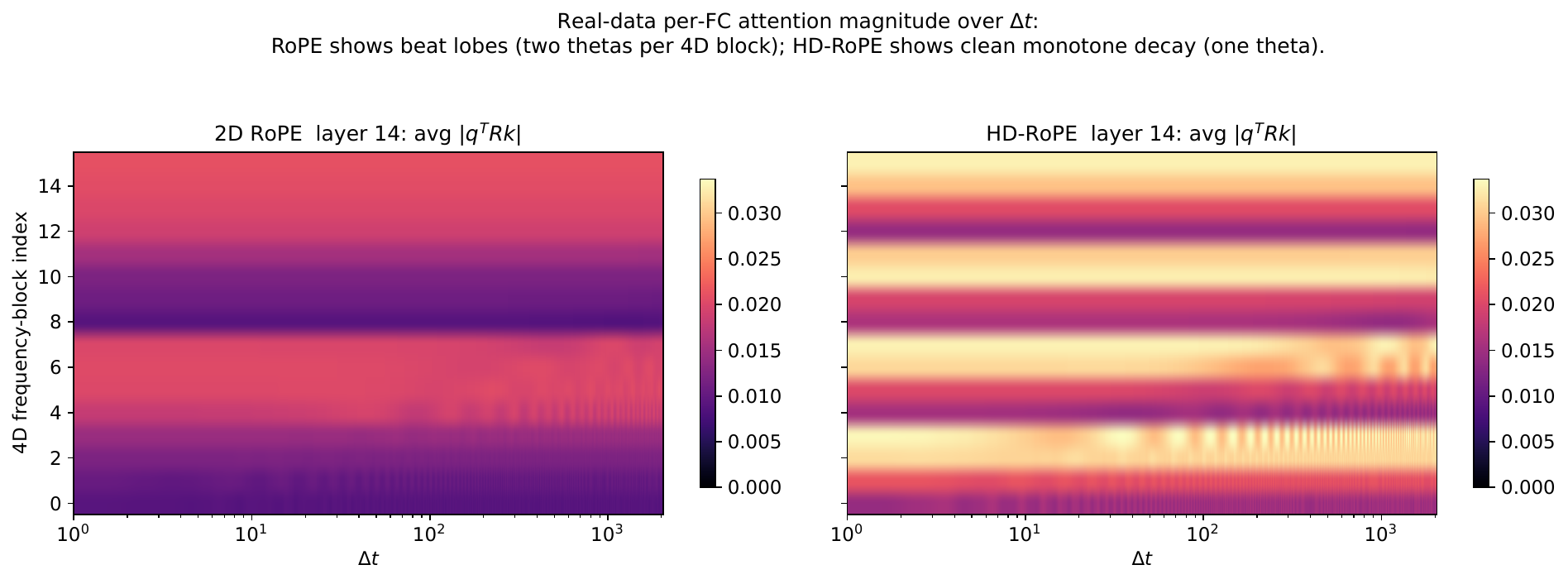}}
    \caption{FC Attention Magnitude of RoPE and HD-RoPE, layer 14. }
    \label{fig:magl0}
  \end{center}
\end{figure}

\begin{figure}[!t]
  \begin{center}
    \centerline{\includegraphics[width=0.95\columnwidth]{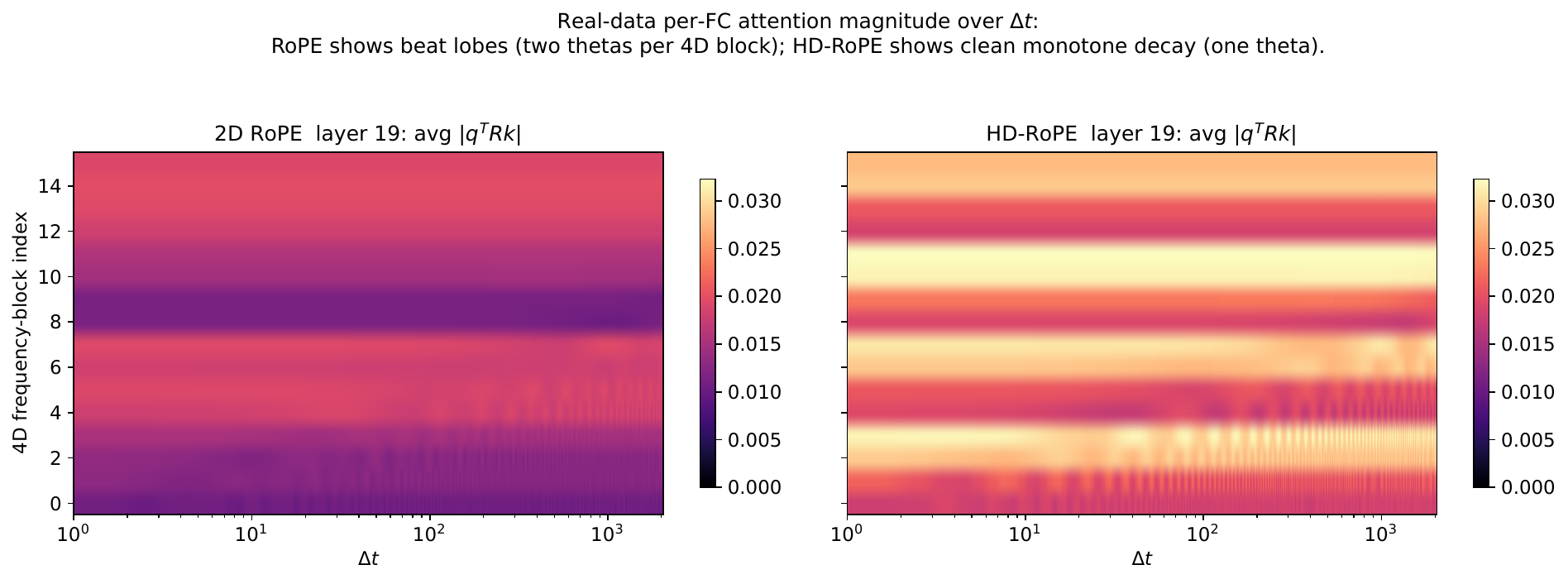}}
    \caption{FC Attention Magnitude of RoPE and HD-RoPE, layer 19. }
    \label{fig:magl0}
  \end{center}
\end{figure}

\subsection{Structured Orthogonal Families for Reproducible Deep Mixing}\label{app:prelim:structured_orthogonal}

A central component of HD-RoPE is a fixed orthogonal mixing transform that re-parameterizes the rotation subspace without altering the norm or the relative-displacement form inherited from RoPE.
While the classical literature offers many orthogonal constructions (e.g., Haar-random orthogonal matrices or QR/Gram--Schmidt generation), such approaches are typically either \emph{random} or \emph{numerically decomposed}, which complicates strict reproducibility across platforms and introduces non-negligible overhead when the transform must be instantiated repeatedly (e.g., across layers/heads or for multiple target dimensions).

\paragraph{Design requirements.}
We restrict attention to a class of orthogonal transforms $\mathbf Q\in\mathbb{R}^{N\times N}$ that satisfy, simultaneously:
\begin{enumerate}[leftmargin=1.2em,itemsep=0.2em,topsep=0.2em]
\item \textbf{Determinism and reproducibility:} $\mathbf Q$ is generated by a deterministic rule given $(N,\text{seed-free})$, avoiding randomness and floating-point decompositions whenever possible.
\item \textbf{Numerical stability:} $\mathbf Q^\top\mathbf Q=\mathbf I$ (exactly or up to a controlled machine-precision error), ensuring $\|\mathbf Q\mathbf x\|_2=\|\mathbf x\|_2$ for all $\mathbf x$.
\item \textbf{Scalability:} the construction extends to many target orders $N$ (ideally an infinite family), or can be composed to cover arbitrary $N$ with minimal artifacts.
\item \textbf{Low overhead:} applying $\mathbf Q$ admits either a fast algorithm (sub-quadratic) or a highly structured implementation (e.g., sign flips and additions) with small constant factors.
\item \textbf{Strong cross-dimensional mixing:} $\mathbf Q$ should distribute each coordinate across many dimensions, rather than acting as a near-permutation.
A convenient quantitative proxy is the \emph{incoherence}
\begin{equation}
\mu(\mathbf Q)\;\triangleq\;\sqrt{N}\,\max_{i,j}|\mathbf Q_{ij}|,
\end{equation}
where smaller $\mu(\mathbf Q)$ indicates more uniform energy spreading from standard basis vectors to mixed coordinates.
\end{enumerate}

\paragraph{Why common orthogonal constructions are insufficient under these constraints.}
Under the above combined constraints, the space of viable orthogonal families becomes notably limited.

\textbf{(1) Haar-random orthogonal and QR/Gram--Schmidt generation.}
Haar-random orthogonal matrices achieve near-optimal isotropy and typically small incoherence with high probability.
However, they rely on randomness and floating-point factorization (e.g., QR), which (i) weakens strict determinism across implementations, and (ii) incurs overhead for generation (typically $\mathcal O(N^3)$) that is undesirable when $N$ varies across experimental settings or must be instantiated repeatedly.
Moreover, numerical decomposition can introduce subtle platform-dependent discrepancies, which is undesirable for reproducible ablation and deployment.

\textbf{(2) Hadamard/Schur-type matrices.}
Hadamard matrices (after normalization) are deterministic and perfectly orthogonal, with entries $\pm 1/\sqrt{N}$ and thus $\mu(\mathbf Q)=1$, which is near-optimal.
They also admit fast transforms (Walsh--Hadamard) with $\mathcal O(N\log N)$ complexity.
Nevertheless, their existence is restricted to specific orders (classically $N=1,2$ or $N\equiv 0\ (\mathrm{mod}\ 4)$, with nontrivial existence gaps in general), limiting their direct applicability when one requires a construction at a prescribed $N$.
In addition, many Hadamard-based transforms are ``too regular'' for certain mixing patterns we target (e.g., enforcing a zero diagonal or other structural constraints), which motivates considering broader design families.

\textbf{(3) Discrete cosine/sine transforms (DCT/DST).}
DCT/DST are deterministic, orthogonal, and have mature fast algorithms.
However, these transforms are intrinsically \emph{harmonic bases} with strong frequency ordering and boundary-condition structure: the mixing they induce is highly anisotropic and biased toward sinusoidal modes.
Consequently, while they are excellent for frequency-domain reparameterization, they do not necessarily realize the kind of near-uniform cross-dimensional coupling desirable for ``deep rotation'' in positional encoding, especially when we aim to avoid injecting additional inductive biases unrelated to position.

\textbf{(4) Sparse orthogonal parameterizations via Givens rotations.}
Products of Givens rotations yield a flexible class of orthogonal transforms.
Yet, achieving strong mixing (e.g., small incoherence and broad support) typically requires a sufficiently deep rotation network, increasing both implementation complexity and runtime cost.
If the rotation angles or rotation pairs are randomized, the determinism requirement is violated; if they are fixed deterministically and kept shallow for efficiency, the resulting transform may remain too sparse/structured and fail to provide the desired mixing strength.

\paragraph{Implication.}
The above discussion indicates that, when we simultaneously require \emph{(i)} deterministic, stable, reproducible generation at prescribed dimensions,
\emph{(ii)} low overhead, and \emph{(iii)} strong cross-dimensional mixing,
the set of suitable structured orthogonal families is comparatively narrow.
This motivates our choice of a highly structured design family:
conference-matrix-derived orthogonal transforms provide a deterministic, discrete, and strongly mixing construction, and are particularly convenient when used in small block sizes (e.g., $N=4,8,32$), where the $\mathcal O(N^2)$ mixing cost is negligible relative to the attention/MLP computation while still yielding substantial representational gains.

\subsection{Additional Background on Conference Matrices}
\label{app:conference}

\subsubsection{Definition and basic constraints}
A (real) conference matrix of order $N$ is a matrix $\mathbf C_N\in\{0,\pm1\}^{N\times N}$
with $\mathrm{diag}(\mathbf C_N)=\mathbf 0$ and
\begin{equation}
\mathbf C_N^\top \mathbf C_N = (N-1)\mathbf I.
\label{eq:conf_def}
\end{equation}
Equation~\eqref{eq:conf_def} implies that the $N$ rows of $\mathbf C_N$ form an orthogonal system
with equal squared norm $(N-1)$.
A direct consequence is that $N$ must be even (for $N>1$), hence conference matrices cannot exist
for arbitrary orders. Moreover, after normalization $\mathbf Q_N=\mathbf C_N/\sqrt{N-1}$,
one obtains an orthogonal transform $\mathbf Q_N^\top\mathbf Q_N=\mathbf I$ used in our deep-mixing
construction. 

\subsubsection{Symmetric vs.\ skew-symmetric types and congruence conditions}
Real conference matrices (up to equivalence operations such as negating rows/columns) fall into two
structured types governed by modular constraints on $N$:
\begin{itemize}[leftmargin=1.2em,itemsep=0.2em,topsep=0.2em]
\item \textbf{Oddly even order ($N\equiv 2\ \mathrm{mod}\ 4$):} the core submatrix is symmetric, yielding
a \emph{symmetric} conference matrix.
\item \textbf{Evenly even order ($N\equiv 0\ \mathrm{mod}\ 4$):} the core submatrix is skew-symmetric, yielding
a \emph{skew-symmetric} conference matrix.
\end{itemize}
In particular, skew-symmetric conference matrices can only occur at orders $N$ divisible by $4$.
These congruence constraints already rule out infinitely many dimensions. 

\subsubsection{Paley-type constructions and admissible orders}
A widely used deterministic infinite family is given by Paley-type constructions based on finite fields.
Let $q$ be an odd prime power and set $N=q+1$. Then:
\begin{itemize}[leftmargin=1.2em,itemsep=0.2em,topsep=0.2em]
\item If $q\equiv 1\ (\mathrm{mod}\ 4)$, Paley graphs yield \emph{symmetric} conference matrices of order $N=q+1$.
\item If $q\equiv 3\ (\mathrm{mod}\ 4)$, Paley digraphs (Paley tournaments) yield \emph{skew-symmetric} conference matrices
of order $N=q+1$, hence $N\equiv 0\ (\mathrm{mod}\ 4)$.
\end{itemize}
Therefore, Paley-type constructions do \emph{not} exist for arbitrary $N$: they only realize orders of the form
$N=q+1$ with $q$ an odd prime power (and with the additional residue condition selecting the symmetric vs.\ skew-symmetric type).

\paragraph{Examples used in this paper.}
Our typical block sizes $N\in\{4,8,32\}$ correspond to $q\in\{3,7,31\}$, all of which are prime powers with
$q\equiv 3\ (\mathrm{mod}\ 4)$. Hence, a skew-symmetric Paley-type conference matrix $\mathbf C_N$ exists for each such $N$,
allowing deterministic and non-random generation of $\mathbf Q_N=\mathbf C_N/\sqrt{N-1}$. 

\subsubsection{Why the existence problem is nontrivial beyond Paley orders}
The defining constraint $\mathbf C_N^\top\mathbf C_N=(N-1)\mathbf I$ is highly rigid:
it requires $N$ sign-vectors with exactly one zero per row/column and pairwise orthogonality.
Even when the congruence conditions above are satisfied, existence is not guaranteed for all admissible $N$,
and the set of known orders is incomplete; Paley-type constructions cover only a subset.
In addition, further arithmetic necessary conditions are known (e.g., constraints on $N-1$),
highlighting that the existence question is governed by deep combinatorial-design and number-theoretic structure.

\subsubsection{Practical handling for arbitrary dimensions}
Since conference matrices do not exist for every $N$, our implementation supports arbitrary target dimensions by composing
available blocks: we construct $\mathbf Q=\mathrm{diag}(\mathbf Q_{N_1},\ldots,\mathbf Q_{N_t},\mathbf I)$ using admissible
orders $\{N_i\}$ (e.g., Paley-type $4,8,32,\ldots$), and pad the remaining dimensions with an identity block.
This preserves orthogonality and enables strong mixing within conference blocks while maintaining deterministic reproducibility.

\subsection{Passkey Retrieval Dataset Construction}\label{sec:passkey_dataset}

We construct a synthetic \emph{Passkey Retrieval} benchmark that matches the LongRoPE-style evaluation setting, with the goal of isolating long-context \emph{needle-in-a-haystack} retrieval under precise token-length control. Each instance is a JSONL record with fields \texttt{\{question, answer\}}, where the \texttt{question} contains a short passkey statement embedded within a large amount of irrelevant text, and the \texttt{answer} is the passkey string.

\paragraph{Tokenizer-consistent length measurement.}
All lengths and constraints are defined in \emph{token space} using the GPT-2 tokenizer (\texttt{use\_fast=True}) with \texttt{add\_special\_tokens=False}. This ensures that (i) context length targets correspond to the model’s effective sequence length, and (ii) dataset construction is reproducible and comparable across runs. We denote by $\tau(\cdot)$ the tokenization operator and by $|\tau(\cdot)|$ the resulting token count.

\paragraph{Prompt template and controllable key position.}
Each sample is generated from four fixed text components:
(i) a short instruction header $H$,
(ii) a filler sentence $F$ repeated many times,
(iii) a key block $K(p)$ that explicitly states the passkey $p$ (repeated to reduce ambiguity),
and (iv) a query suffix $Q$ that asks for the passkey. Concretely, for integers $x,y\ge 1$, we form
\begin{equation}
\texttt{question}(x,y,p) \;=\; H \;\Vert\; F^{x} \;\Vert\; K(p) \;\Vert\; F^{y} \;\Vert\; Q,
\end{equation}
where $\Vert$ denotes string concatenation. The pair $(x,y)$ controls the passkey’s location: $x$ determines the distance from the beginning (prefix filler), and $y$ determines the distance from the end (suffix filler before the query). This design decouples \emph{context length} from \emph{key position}, enabling systematic coverage of retrieval difficulty patterns.

\paragraph{Single-token passkey pool.}
To make evaluation unambiguous at the token level, passkeys are restricted to digit strings that tokenize into a \emph{single token} under GPT-2 tokenization. By default, we enforce that the \emph{leading-space context} (\texttt{" " + digits}) is a single token, matching the typical in-context occurrence after natural language. Optionally, we can additionally require the no-space form (\texttt{digits}) to be single-token. We scan a user-specified integer range $[v_{\min}, v_{\max}]$, collect all digit strings satisfying the constraint, and cache the resulting pool (with metadata such as range, tokenizer name, and constraints) to ensure reproducibility and to avoid repeated scanning. During generation, each instance samples $p$ uniformly from this pool with a fixed RNG seed.

\paragraph{Length bucketing and target selection.}
To evaluate over a wide context range while maintaining exact token-based control, we partition samples into token-length buckets of the form $(L, U]$, where $U$ is an upper bound (default width 256 tokens). For very long contexts, we optionally use an \emph{adaptive stride} to reduce the number of buckets while preserving coverage at short lengths. For each bucket, we select a small set of target lengths $\{T^{(j)}\}$ within $(L,U]$ (the number of targets decreases as $U$ grows) and search for an integer sum $s=x+y$ such that the realized length
\begin{equation}
T(x,y,p) = |\tau(\texttt{question}(x,y,p))|
\end{equation}
falls inside the bucket. Because tokenization is not perfectly linear in characters, we compute $T$ directly by tokenizing a probe instance and search over $s$ in a small neighborhood around a linear estimate.

\paragraph{Coverage over $(x,y)$ without exhaustive enumeration.}
Enumerating all $(x,y)$ pairs is unnecessary and expensive. Instead, for each bucket and each feasible $s$, we sample $x$ values using an integer linspace over $[1, s-1]$ with a bucket-dependent number of steps. For each sampled $x$, we set $y=s-x$. This yields monotonic coverage from early-key (small $x$, large $y$) to late-key (large $x$, small $y$) placements at approximately fixed total length, providing broad coverage of retrieval positions while keeping dataset size bounded.

\paragraph{Output organization and size budgeting.}
Each bucket has its own directory named by the bucket upper bound $U$. Within a bucket, each $(x,y)$ setting is written to a dedicated JSONL file whose filename encodes $(x,y)$, their approximate token contributions $(x\cdot |\tau(F)|,\, y\cdot |\tau(F)|)$, and the realized total length $T$. To bound storage and generation cost, we enforce both (i) a global byte budget (default $\sim$1\,GiB) and (ii) per-bucket byte caps that \emph{decrease} with $U$, allocating more samples to shorter contexts and fewer to extremely long contexts. We conservatively estimate bytes per line using a probe passkey with maximal digit length. Generation stops early once either the global or per-bucket budget is exhausted.

\paragraph{Reproducibility and provenance.}
We always emit a \texttt{dataset\_meta.json} file recording the tokenizer, bucket configuration, passkey constraints and range, filler/base token counts, RNG seed, and byte budgets. Optionally, a \texttt{summary.json} aggregates per-bucket counts (files, lines, bytes) and stop reasons. A \texttt{--dry\_run} mode performs the complete planning and budget accounting without writing JSONL files, enabling deterministic capacity planning before committing storage.

Overall, this construction follows three principles: (i) \emph{token-faithful} control of sequence length and key position; (ii) \emph{evaluation clarity} via single-token answers; and (iii) \emph{scalable coverage} across long contexts through adaptive bucketing and structured sampling over $(x,y)$ under strict size budgets.

\end{document}